\documentclass[10pt,twocolumn,letterpaper]{article}

\usepackage{iccv}
\usepackage{times}
\usepackage{epsfig}
\usepackage{graphicx}
\usepackage{amsmath}
\usepackage{amssymb}
\usepackage{graphicx}
\usepackage{epstopdf}
\usepackage{makecell}
\usepackage{multirow}
\usepackage{array}
\usepackage{bm}
\usepackage{dsfont}
\usepackage{color}
\usepackage{caption}
\usepackage{algorithm} 
\usepackage{algorithmic} 
\usepackage{graphicx}
\usepackage{subfigure}
\usepackage{xcolor}

\usepackage{pifont}

\newcommand{\tabincell}[2]{\begin{tabular}{@{}#1@{}}#2\end{tabular}}

\usepackage{enumitem}

\usepackage[pagebackref=true,breaklinks=true,letterpaper=true,colorlinks,bookmarks=false]{hyperref}

\iccvfinalcopy 

\def\iccvPaperID{} 

\ificcvfinal\fi

\begin{document}
\captionsetup{font=footnotesize}
\title{Masked Face Recognition Challenge: The WebFace260M Track Report}

\author{Zheng Zhu\textsuperscript{\rm 1} ~ Guan Huang\textsuperscript{\rm 2} ~ Jiankang Deng\textsuperscript{\rm 3} ~Yun Ye\textsuperscript{\rm 2} ~Junjie Huang\textsuperscript{\rm 2}  ~Xinze Chen\textsuperscript{\rm 2} \\ ~Jiagang Zhu\textsuperscript{\rm 2} ~ Tian Yang\textsuperscript{\rm 2} ~Jia Guo\textsuperscript{\rm 4} ~Jiwen Lu\textsuperscript{\rm 1} ~Dalong Du\textsuperscript{\rm 2} ~Jie Zhou\textsuperscript{\rm 1}\\
\textsuperscript{\rm 1}Tsinghua University
~ ~ \textsuperscript{\rm 2}XForwardAI
~ ~ \textsuperscript{\rm 3}Imperial College London
~ ~ \textsuperscript{\rm 4}InsightFace
\\
\tt\small \{zhengzhu,lujiwen\}@tsinghua.edu.cn    \{guan.huang,dalong.du\}@xforwardai.com \\
\tt\small j.deng16@imperial.ac.uk
}

\maketitle
\ificcvfinal\thispagestyle{empty}\fi

\begin{abstract}

According to WHO statistics, there are more than 204,617,027 confirmed COVID-19 cases including 4,323,247 deaths worldwide till August 12, 2021. During the coronavirus epidemic, almost everyone wears a facial mask. Traditionally, face recognition approaches process mostly non-occluded faces, which include primary facial features such as the eyes, nose, and mouth.
Removing the mask for authentication in airports or laboratories will increase the risk of virus infection, posing a huge challenge to current face recognition systems. Due to the sudden outbreak of the
epidemic, there are yet no publicly available real-world masked face
recognition (MFR) benchmark.
To cope with the above-mentioned issue, we organize the Face Bio-metrics under COVID Workshop and Masked Face Recognition Challenge in ICCV 2021.
Enabled by the ultra-large-scale WebFace260M benchmark and the Face Recognition Under Inference Time conStraint (FRUITS) protocol, this challenge (WebFace260M Track) aims to push the frontiers of practical MFR.
Since public evaluation sets are mostly saturated or contain noise, a new test set is gathered consisting of elaborated 2,478 celebrities and 60,926 faces.
Meanwhile, we collect the world-largest real-world masked test set.
In the first phase of WebFace260M Track, 69 teams (total 833 solutions) participate in the challenge and 49 teams exceed the performance of our baseline.
There are second phase of the challenge till October 1, 2021 and on-going leaderboard.
We will actively update this report in the future.

\end{abstract}

\vspace{-5mm}

\section{Introduction}

Due to the boom of CNNs, standard face recognition (SFR) systems have achieved a remarkable success, which usually work with mostly non-occluded faces. However, there are a number of circumstances where faces are occluded by facial masks, rising the masked face recognition (MFR) problem.

During the global COVID-19, people are encouraged to wear masks in public areas, making primary facial features invisible. Few SFR systems can work well with this situation, but removing the mask for authentication will increase the risk of virus infection. Recently, some commercial vendors \cite{FRVT-mask} have developed face recognition algorithms capable of handling face masks, and an increasing number of research publications \cite{ding2020masked,geng2020masked,du2021towards,hariri2021efficient,anwar2020masked} have surfaced on this topic. However, due to the sudden outbreak of the epidemic, there are yet no publicly available large-scale MFR benchmark.

To address the above-mentioned issue, we organize the Face Bio-metrics under COVID Workshop and Masked Face Recognition Challenge in ICCV 2021. Face benchmarks empower researchers to train high-performance face recognition systems.
Enabled by the ultra-large-scale WebFace260M benchmark \cite{WebFace260M}, this challenge aims to push the frontiers of practical MFR.
On the other hand, evaluation protocols and test set play an essential role in analysing face recognition performance.
Since public evaluation sets are mostly saturated or contain noise, we adopt the Face Recognition Under Inference Time conStraint (FRUITS) protocol in WebFace260M Track in this workshop.
Besides, a new test set is gathered consisting of elaborated 2,478 celebrities and 60,926 faces.
Meanwhile, we collect the world-largest real-world masked test set.

This paper is the official report of WebFace260M Track in the MFR workshop and challenge.
We detail the training data, evaluation protocols, submission rules, test set and metric in SFR and MFR, ranking criterion, baseline solution, and preliminary competition results.
The challenge was launched at June 7, 2021.
In the first phase of WebFace260M Track, 69 teams from academia and industry participate in the challenge and 49 teams exceed the performance of our baseline. Total 833 solutions are submitted, covering various network designs and training strategies.
There are second phase of the challenge till October 1, 2021 and on-going leaderboard.
We will actively update this report in the future.

\begin{table*}[!t]
\begin{center}{\scalebox{0.9}{
\begin{tabular}{l|c|c|c|c|c|c|c}
\hline
Dataset & \# Identities & \# Images & Images/ID  & Cleaning  & \# Attributes & Availability & Publications\\ \hline
\hline
CASIA-WebFace~\cite{CASIA-WebFace} & $10$ K & $0.5$ M & 47 & Auto & - & Public & Arxiv 2014\\
CelebFaces~\cite{DeepID} & $10$ K & $0.2$ M & 20 & Manual & 40  & Public & ICCV 2015\\
UMDFaces~\cite{UMDFaces} & $8$ K & $0.3$ M & 45 & Semi-auto & 4 & Public & IJCB 2017\\
VGGFace~\cite{VGGFace} & $2$ K & $2.6$ M & 1,000 & Semi-auto & - & Public & BMVC 2015\\
VGGFace2~\cite{VGGFace2} & $9$ K & $3.3$ M & 363 & Semi-auto & 11 & Public & FG 2018\\
MS1M~\cite{MS1M} &$0.1$ M & $10$ M & 100 & No & -& Public & ECCV 2016\\
MS1M-IBUG~\cite{deng2017marginal} &$85$ K & $3.8$ M & 45 & Semi-auto & -& Public & CVPRW 2017\\
MS1MV2~\cite{ArcFace} & $85$ K & $5.8$ M & 68 & Semi-auto & -& Public & CVPR 2019\\
MS1M-Glint~\cite{glintweb} & $87$ K & $3.9$ M  & 44 & Semi-auto & -& Public & -\\
MegaFace2~\cite{MF2} & $0.6$ M & $4.7$ M & 7 & Auto & -& Public & CVPR 2017\\
IMDB-Face~\cite{IMDB-Face} & $59$ K & $1.7$ M & 29 & Manual &  - & Public & ECCV 2018\\ \hline
\textcolor[RGB]{128,128,128}{Facebook~\cite{DeepFace}} & \textcolor[RGB]{128,128,128}{$4$ K}  & \textcolor[RGB]{128,128,128}{$4.4$ M} & \textcolor[RGB]{128,128,128}{1,100} & \textcolor[RGB]{128,128,128}{-} & \textcolor[RGB]{128,128,128}{-}& \textcolor[RGB]{128,128,128}{Private} &\textcolor[RGB]{128,128,128}{CVPR 2014}\\
\textcolor[RGB]{128,128,128}{Facebook~\cite{taigman2015web}} & \textcolor[RGB]{128,128,128}{$10$ M} & \textcolor[RGB]{128,128,128}{$500$ M} & \textcolor[RGB]{128,128,128}{50} & \textcolor[RGB]{128,128,128}{-} & \textcolor[RGB]{128,128,128}{-}& \textcolor[RGB]{128,128,128}{Private} &\textcolor[RGB]{128,128,128}{CVPR 2015}\\
\textcolor[RGB]{128,128,128}{Google~\cite{FaceNet}} & \textcolor[RGB]{128,128,128}{$8$ M} & \textcolor[RGB]{128,128,128}{$200$ M} & \textcolor[RGB]{128,128,128}{25} & \textcolor[RGB]{128,128,128}{-} & \textcolor[RGB]{128,128,128}{-}& \textcolor[RGB]{128,128,128}{Private} & \textcolor[RGB]{128,128,128}{CVPR 2015}\\
\textcolor[RGB]{128,128,128}{MillionCelebs~\cite{MillionCelebs}} & \textcolor[RGB]{128,128,128}{$0.6$ M} & \textcolor[RGB]{128,128,128}{$18.8$ M} & \textcolor[RGB]{128,128,128}{30} & \textcolor[RGB]{128,128,128}{Auto} & \textcolor[RGB]{128,128,128}{-}& \textcolor[RGB]{128,128,128}{Private} & \textcolor[RGB]{128,128,128}{CVPR 2020}\\ \hline
\textbf{WebFace260M}  & \textbf{4 M}  & \textbf{260M}  & \textbf{65} & No & -& Public & CVPR 2021 \\
\textbf{WebFace42M} & \textbf{2 M}  & \textbf{42M}  & \textbf{21} & Auto & 7 & Public & CVPR 2021 \\ \hline
\end{tabular}}}
\end{center}
\caption{Training data for deep face recognition. The cleaned WebFace42M is the largest public training set in terms of both \# identities and \# images.}
\label{table:training_set}
\end{table*}

\section{Training Data and Evaluation Protocols}

\subsection{WebFace260M Data}

WebFace260M \cite{WebFace260M} is current largest public face recognition dataset, covering noisy 4M identities/260M faces and cleaned 2M identities/42M faces. With such large data size, this dataset takes a significant step towards closing the data gap between academia and industry as shown in Table~\ref{table:training_set}. The celebrity name list consists of two parts: the first one is borrowed from MS1M (1 million, constructed from Freebase) and the second one (3 millions) is collected from the IMDB database. Based on the name list, celebrity faces are searched and downloaded via Google image search engine~\cite{google_image}. The WebFace42M training set is obtained by a Cleaning Automatically utilizing Self-Training (CAST) pipeline.
Noise ratio of WebFace42M is lower than 10\% (similar to CASIA-WebFace~\cite{CASIA-WebFace} and Glint360K \cite{an2020partial}) based on the sampling estimation.
After CAST, duplicates of each subject are removed when their cosine similarity is higher than 0.95. Furthermore, the feature center of each subject is compared with popular benchmarks (\eg the test set in this challenge, LFW families~\cite{LFW,CALFW,CPLFW}, FaceScrub~\cite{FaceScrub}, IJB-C~\cite{IJB-C} \etal), and overlaps are removed if the cosine similarity is higher than 0.7.


\subsection{FRUITS Protocol}

Most existing face recognition evaluation protocols \cite{LFW,CFP,AgeDB,CALFW,CPLFW,MegaFace,IJB-C,YTF,IQIYI2018} target the pursuit of accuracy. However, face recognition in real-world application scenarios is always restricted by inference time.
Lightweight face recognition challenge \cite{LFR} takes a step toward this goal by constraining the FLOPs and model size of submissions, which is not a straightforward solution. Besides, it neglects the face detection and alignment module cost.
Strict submission policy of NIST-FRVT~\cite{FRVT} hinders researchers to freely evaluate their algorithms.
In WebFace260M Track of this challenge, we follow the Face Recognition Under Inference Time conStraint (FRUITS) protocol. Referring to~\cite{WebFace260M},  inference time is measured on a single core of an Intel Xeon CPU E5-2630-v4@2.20GHz processor (no GPU hardware is provided), and the constraint of 1000 milliseconds is adopted.

\begin{figure*}[t]
\centering
\includegraphics[width=0.9\linewidth]{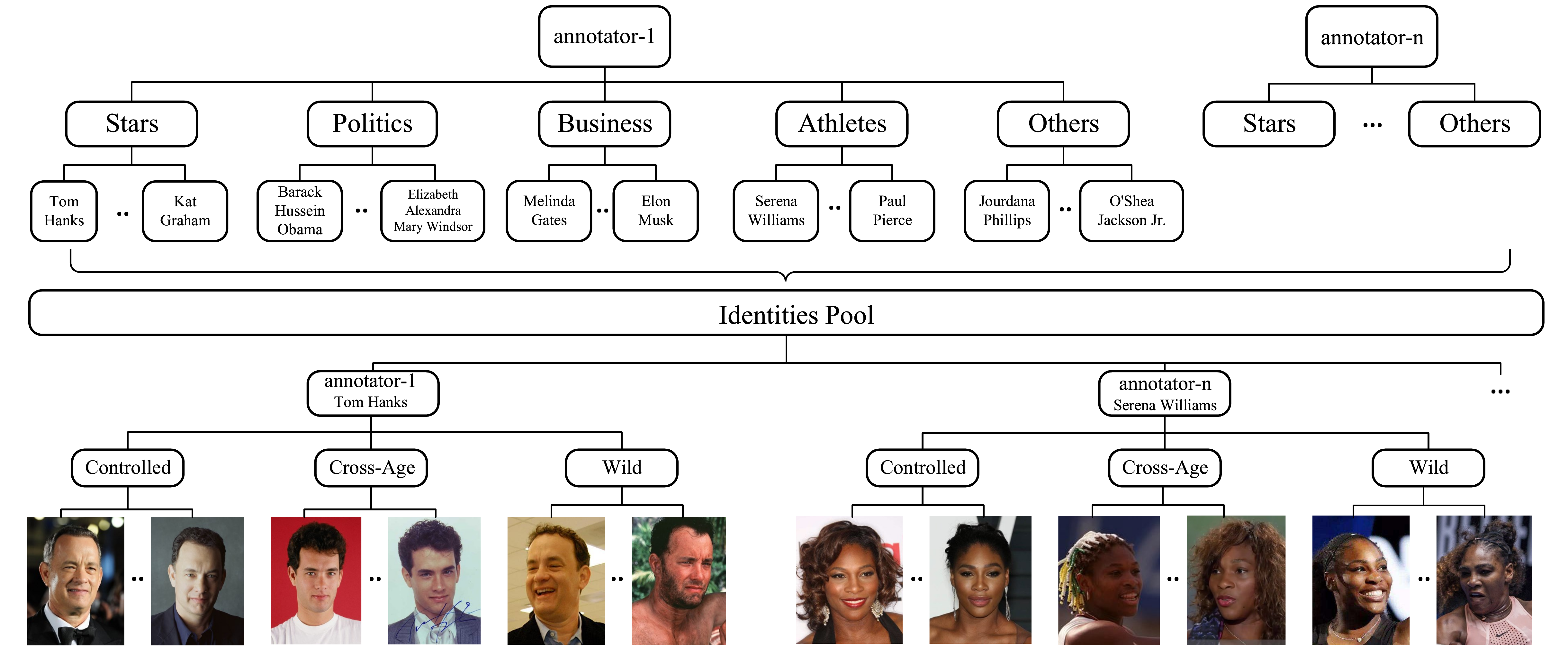}
   \caption{The collecting pipeline of test set.}
  \vspace{-2mm}
\label{fig:test_set}
\end{figure*}

\subsection{Submission Rules}

The WebFace260M Track of our challenge\footnote{\url{https://www.face-benchmark.org/challenge.html}} has two phases.
For the first phase, the number of max submissions per day is 5.
For the second phase, the number of max submissions per day is 3.
The full WebFace260M data has been open for all applicants, as long as their agreements\footnote{\url{https://www.face-benchmark.org/doc/license_agreement_for_webface260m_dataset.pdf}} are qualified. Mask data-augmentation is allowed, for example this method\footnote{\url{https://github.com/deepinsight/insightface/tree/master/recognition/_tools_}}. The applied mask augmentation tool should be reproducible. External dataset and pre-trained models are both prohibited in the second phase. Participants could submit their
submission package\footnote{\url{https://github.com/WebFace260M/webface260m-iccv21-mfr}} to the submission server\footnote{\url{https://competitions.codalab.org/competitions/32478}} and get scores by our online evaluation.
Participants should run the code for verification on the provided docker file to ensure the correctness of feature and time constraints. Both the models for face detection and recognition should be converted to ONNX format.
Participants must package the code directory for submission and see the results on the leaderboard\footnote{\url{https://competitions.codalab.org/competitions/32478\#results}}.
Test images are invisible during challenges.

\section{Standard Face Recognition}

\subsection{Test Set}

\begin{figure}[t]
\centering
\includegraphics[width=1.0\linewidth]{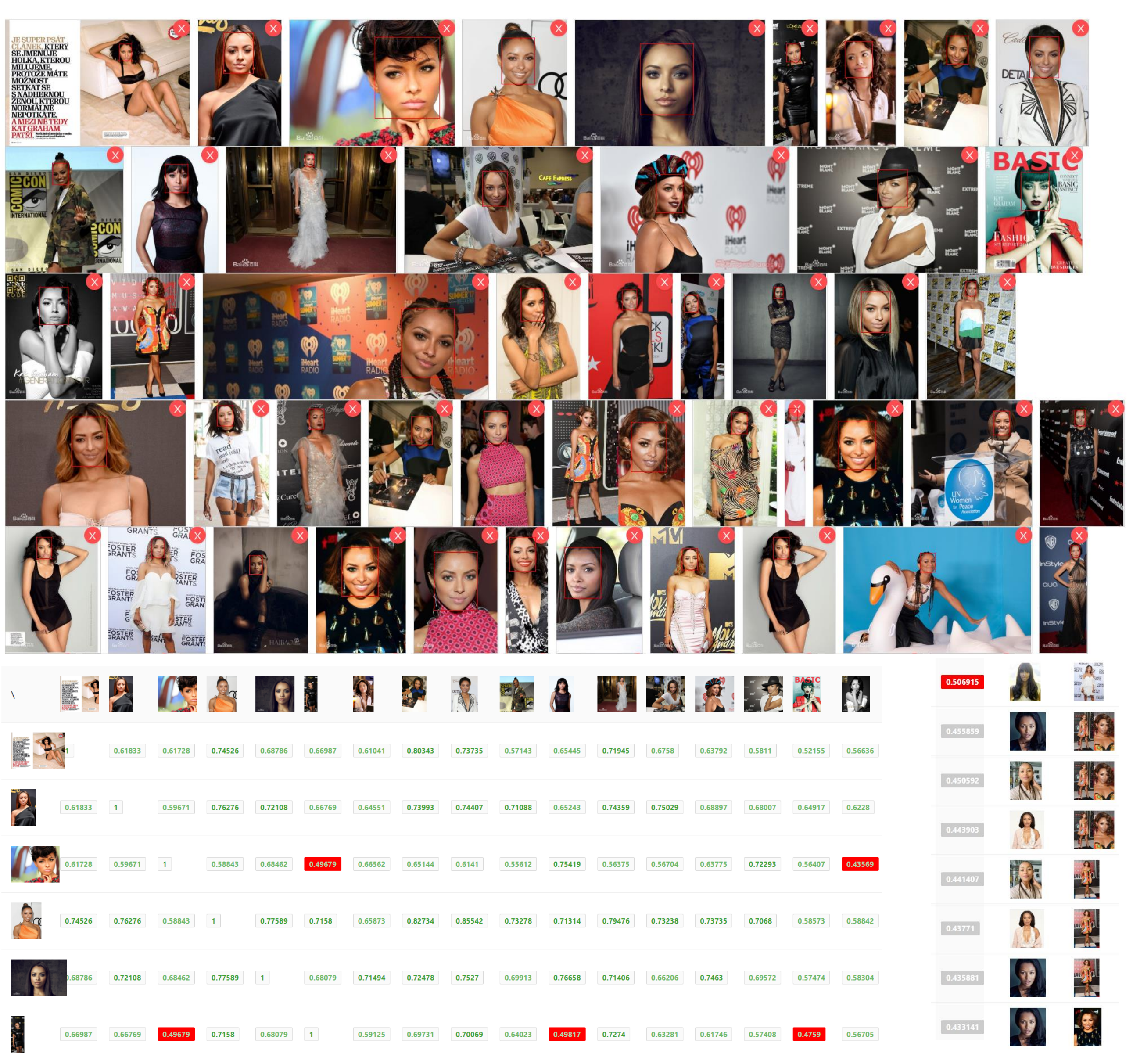}
   \caption{The annotation interface of our test set collecting system. Top part is gathered celebrities images, while bottom part shows cosine similarities.}
\label{fig:system}
\vspace{-5mm}
\end{figure}

To compare deep CNN face matchers utilizing FRUITS protocol, we manually construct a elaborated test set. It is well known that recognizing strangers (especially when they are similar-looking) is a difficult task even for experienced researchers. So we choose to select our familiar celebrities, which ensure the high-quality of the test set.

The collecting pipeline of test set is illustrated in Figure \ref{fig:test_set}. Each annotator is asked to write about 100 names of celebrities including stars, politics, business, athletes \etal. Different name lists are exclusive, and a identities pool is maintained by merging lists. It is noting that annotators are encouraged to collect gender-balanced and race-balanced identities lists.

The final identities pool consists of 2,748 names. For each celebrity, responsible annotator need to collect his/her faces from reliable sources of information. Targeting at analysing recognition performance in different application scenarios,
3 subclass is defined to guide collection:

\noindent{\bf Controlled image}: \emph{Controlled} faces collection targets to evaluate ID photo such as visa and driving license. \emph{Controlled} face in our test set is defined as: near frontal, five landmarks are visible, normal expression, not low-resolution.


\noindent{\bf Wild image}: \emph{Wild} subclass aims to collect faces in unconstrained scenarios, including large pose, partial occlusion, resolution variation, illumination variation \etal.

\noindent{\bf Cross-age image}: This subclass collects faces whose age is obviously different from \emph{Controlled} and \emph{Wild} faces, including \emph{Cross-age-10} (more than 10-years gap) and \emph{Cross-age-20} (more than 20-year gap).

For each subclass of a certain celebrity, we gather about 7 faces (\ie ~20 faces/identity) to construct test set as shown in Figure \ref{fig:test_set}. Furthermore, we design a interface to assist annotators to judge the difficulty and quality of the collected faces. In Figure \ref{fig:system}, collected images for a celebrity, its intra-class scores, its highest inter-class scores with other identities are illustrated respectively. Based on scores  indicator from this system, annotators are encouraged to gather hard case (For example, less than 0.5 similarity score for intra-class, more than 0.5 for inter-class). Besides, noisy faces could be effectively filtered (For example, if the score of one face is less than 0.5 compared with all other intra-class images, it should be paid more attention).

The statistics of final test is listed in Table \ref{table:test_set_num}. There are elaborately constructed 2,478 identities and 57,715 faces in total for SFR. 3,328,950,920 impostors and 2,070,305 genuine pairs could be constructed.
Protected attributes (gender, race) as well as different scenarios (\emph{Controlled}, \emph{Wild}, \emph{Cross-age}) are accurately annotated for each subject. \emph{Cross-age} and \emph{Cross-scene} comparisons are also conducted in corresponding subset.

\begin{table}[t]
\begin{center}{\scalebox{0.67}{
\begin{tabular}{c|c|c|c|c|c}
\hline
Eva. &
Attributes& \# Identities & \# Faces & \# Impostor & \# Genuine \\ \hline
\hline

\multirow{6}{*}{\tabincell{l}{SFR}}

&\textbf{All} & \textbf{2,478} & \textbf{57,715} & \textbf{3,328,950,920} & \textbf{2,070,305}\\
\cline{2-6}

&Cross-age-10& - & - & 1,667,036,254 & 553,174\\
&Cross-age-20& - & - & 842,366,636 & 123,560 \\ \cline{2-6}

& Controlled& - & 22,135 & 489,657,590  & 300,635\\
&Wild& - &  35,580   & 1,264,898,172 & 1,038,228\\
&Cross-scene& - & - & 1,574,395,158 & 731,442   \\  \hline

\multirow{6}{*}{\tabincell{l}{MFR}}

&
\textbf{All} & \textbf{2,478} & \textbf{60,926} & - & -\\
\cline{2-6}

&

Masked & 862  & 3,211 & - & -\\
\cline{2-6}

&

Nonmasked & 2,478 & 57,715 & - & -\\
\cline{2-6}

& Controlled-Masked& - & - & 71,042,982  & 32,503  \\
&Wild-Masked& - &  -   & 114,193,476 & 53,904\\
&All-Masked& - & - & 185,236,458 & 86,407   \\  \hline

%

\end{tabular}}}
\end{center}
\vspace{-4mm}
\caption{The statistics of our test set.}
\label{table:test_set_num}
\end{table}


\begin{figure}
\small
\centering
\includegraphics[width=0.83\linewidth]{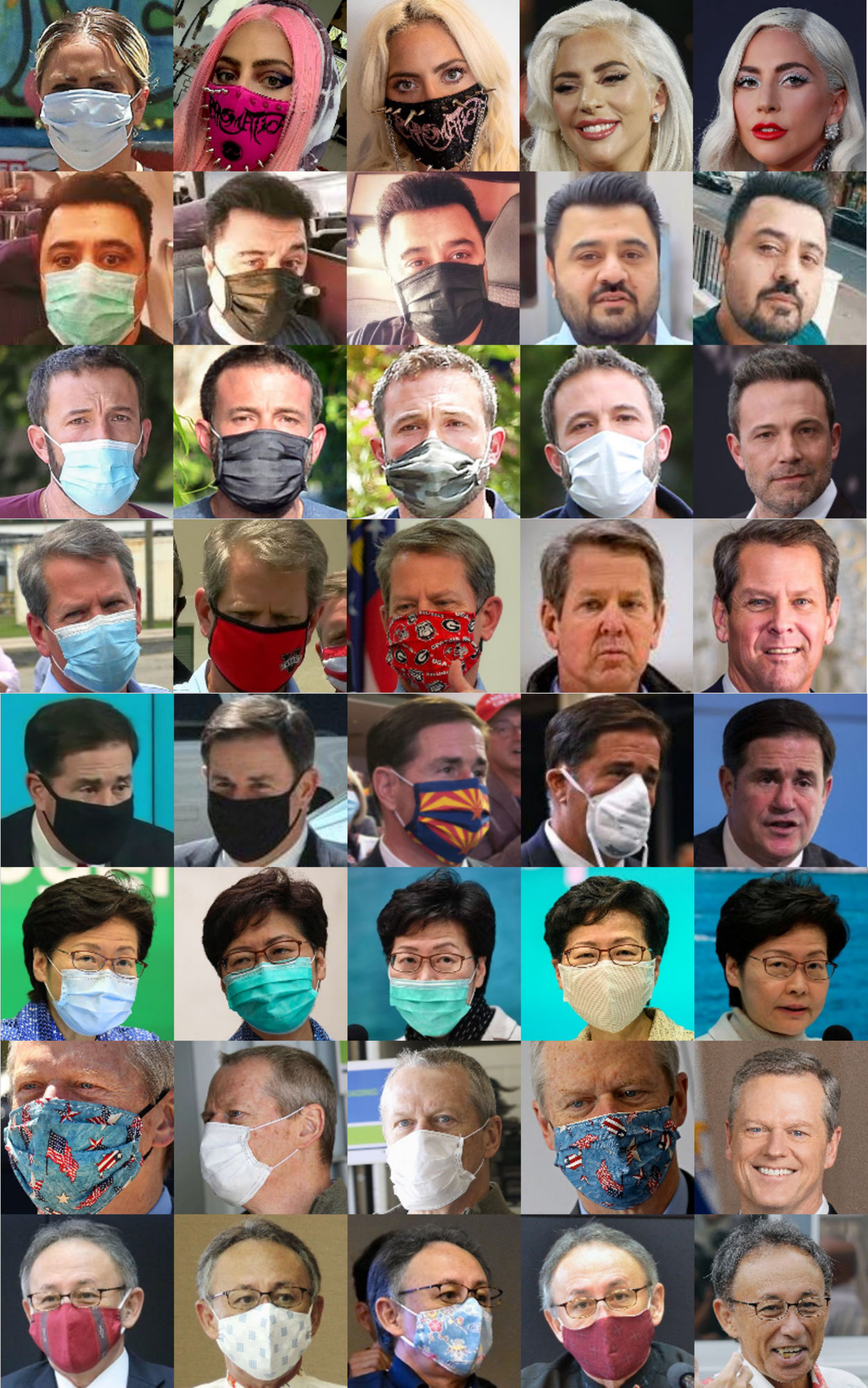}
\caption{Celebrities with and without real-world mask.}
\vspace{-5mm}
\label{fig:mask_test_set}
\end{figure}

\subsection{Metric}

Based on the FRUITS protocol and the new test set, we perform 1:1 face verification across various attributes for SFR evaluation.  Table \ref{table:test_set_num} shows numbers of imposter and genuine in different verification settings. \emph{All} means impostors are paired without attention to any attribute, while later comparisons are conducted on age and scenario subsets. \emph{Cross-age} refers to cross-age (more than 10 and 20 years) verification, while \emph{Cross-scene} means pairs are compared between controlled and wild settings. Different algorithms are measured on False Non-Match Rate (FNMR) \cite{FRVT}, which is defined as the proportion of mated comparisons below a threshold set to achieve the False Match Rate (FMR) specified. FMR is the proportion of impostor comparisons at or above that threshold. It is worth noting that \textbf{Lower FNMR at the same FMR is better}.


\section{Masked Face Recognition}


\subsection{Test Set}

In contrast with simulated \cite{FRVT-mask, negi2021deep} or relatively small \cite{anwar2020masked,damer2021extended,IJCB-mask,wang2020masked} masked face test sets, a real-world comprehensive benchmark for evaluating MFR is developed in this challenge. Based on the SFR identities, we further collect masked faces for these celebrities. Specifically, as shown in Table \ref{table:test_set_num}, there are carefully selected 3,211 masked faces among 862 identities. Subjects with real-world mask are illustrated in Figure~\ref{fig:mask_test_set}.

\subsection{Metric}

For MFR, assessment is performed with \emph{Mask-Nonmask} comparisons. Specifically, there is one face with mask in imposter and genuine, while another face is from standard face sets. According to the attribute of face without mask, we evaluate the performance of algorithms under \emph{Controlled-Masked}, \emph{Wild-Masked}, and \emph{All-Masked} settings listed in Table \ref{table:test_set_num}.

\begin{table*}[!t]
\begin{center}{\scalebox{0.77}{
\begin{tabular}{l|l|c|c|c|c|c|c|c|c|c}
\hline
Rank & Participant & \makecell[c]{All\\\textbf{(MFR\&SFR)}}  & \makecell[c]{Wild\\\textbf{(MFR\&SFR)}}   & \makecell[c]{Controlled\\\textbf{(MFR\&SFR)}}  &  \makecell[c]{All\\\textbf{(SFR)}}  & \makecell[c]{Wild\\\textbf{(SFR)}}   & \makecell[c]{Controlled\\\textbf{(SFR)}}  & \makecell[c]{Detection\\time} & \makecell[c]{Recognition\\time}  & \makecell[c]{Total\\time}                                      \\ \hline
\hline
1 & Ethan.y & 0.0980$^{(1)}$ & 0.1283$^{(5)}$ & 0.0500$^{(1)}$ & 0.0393$^{(12)}$ & 0.0627$^{(13)}$ & 0.0032$^{(6)}$  & 156 & 760 & 916\\
2 & victor-2021 & 0.1017$^{(2)}$ & 0.1222$^{(1)}$ & 0.0694$^{(3)}$ & 0.0162$^{(1)}$ & 0.0270$^{(1)}$ & 0.0018$^{(1)}$  & 157 & 496 & 653\\
3 & sleepybear & 0.1036$^{(3)}$ & 0.1246$^{(2)}$ & 0.0699$^{(5)}$ & 0.0168$^{(2)}$ & 0.0276$^{(2)}$ & 0.0020$^{(2)}$  & 162 & 498 & 660\\
4 & wjtan99 & 0.1056$^{(4)}$ & 0.1279$^{(4)}$ & 0.0698$^{(4)}$ & 0.0187$^{(3)}$ & 0.0306$^{(4)}$ & 0.0021$^{(3)}$  & 97 & 897 & 994\\
5 & hukangli & 0.1056$^{(4)}$ & 0.1278$^{(3)}$ & 0.0698$^{(4)}$ & 0.0187$^{(3)}$ & 0.0305$^{(3)}$ & 0.0021$^{(3)}$  & 97 & 899 & 996\\
6 & min.yang & 0.1131$^{(5)}$ & 0.1366$^{(6)}$ & 0.0758$^{(6)}$ & 0.0228$^{(4)}$ & 0.0367$^{(5)}$ & 0.0026$^{(4)}$  & 158 & 453 & 611\\
7 & wzw & 0.1272$^{(6)}$ & 0.1530$^{(7)}$ & 0.0833$^{(11)}$ & 0.0267$^{(7)}$ & 0.0432$^{(8)}$ & 0.0028$^{(5)}$  & 49 & 744 & 793\\
8 & vuvko & 0.1315$^{(7)}$ & 0.1575$^{(8)}$ & 0.0872$^{(12)}$ & 0.0248$^{(6)}$ & 0.0407$^{(7)}$ & 0.0028$^{(5)}$  & 157 & 926 & 1083\\
9 & lcx2 & 0.1318$^{(8)}$ & 0.1699$^{(13)}$ & 0.0689$^{(2)}$ & 0.0585$^{(27)}$ & 0.0911$^{(30)}$ & 0.0061$^{(19)}$  & 186 & 371 & 557\\
10 & linkpal2021 & 0.1319$^{(9)}$ & 0.1622$^{(11)}$ & 0.0828$^{(10)}$ & 0.0346$^{(9)}$ & 0.0561$^{(10)}$ & 0.0033$^{(7)}$  & 162 & 769 & 931\\
11 & tuolaji & 0.1340$^{(10)}$ & 0.1610$^{(10)}$ & 0.0909$^{(15)}$ & 0.0242$^{(5)}$ & 0.0399$^{(6)}$ & 0.0026$^{(4)}$  & 167 & 286 & 453\\
12 & crishawy & 0.1340$^{(10)}$ & 0.1610$^{(10)}$ & 0.0909$^{(15)}$ & 0.0242$^{(5)}$ & 0.0399$^{(6)}$ & 0.0026$^{(4)}$  & 158 & 861 & 1019\\
13 & betterone & 0.1341$^{(11)}$ & 0.1608$^{(9)}$ & 0.0882$^{(13)}$ & 0.0291$^{(8)}$ & 0.0476$^{(9)}$ & 0.0032$^{(6)}$  & 157 & 948 & 1105\\
14 & cheng3qing & 0.1377$^{(12)}$ & 0.1685$^{(12)}$ & 0.0812$^{(8)}$ & 0.0538$^{(24)}$ & 0.0782$^{(24)}$ & 0.0062$^{(20)}$  & 159 & 688 & 847\\
15 & nayesoj & 0.1389$^{(13)}$ & 0.1724$^{(14)}$ & 0.0814$^{(9)}$ & 0.0498$^{(19)}$ & 0.0761$^{(23)}$ & 0.0042$^{(11)}$  & 164 & 346 & 510\\
16 & amyburden & 0.1503$^{(14)}$ & 0.1807$^{(15)}$ & 0.0996$^{(17)}$ & 0.0413$^{(13)}$ & 0.0639$^{(14)}$ & 0.0048$^{(16)}$  & 158 & 659 & 817\\
17 & mind\_ft & 0.1503$^{(14)}$ & 0.1813$^{(16)}$ & 0.1003$^{(18)}$ & 0.0377$^{(10)}$ & 0.0603$^{(11)}$ & 0.0040$^{(9)}$  & 160 & 340 & 500\\
18 & jyf & 0.1591$^{(15)}$ & 0.2026$^{(19)}$ & 0.0769$^{(7)}$ & 0.0878$^{(41)}$ & 0.1267$^{(42)}$ & 0.0097$^{(30)}$  & 158 & 320 & 478\\
19 & Wison & 0.1596$^{(16)}$ & 0.1906$^{(17)}$ & 0.1093$^{(20)}$ & 0.0386$^{(11)}$ & 0.0624$^{(12)}$ & 0.0037$^{(8)}$  & 164 & 570 & 734\\
20 & ncvl01 & 0.1685$^{(17)}$ & 0.1996$^{(18)}$ & 0.1133$^{(21)}$ & 0.0457$^{(15)}$ & 0.0683$^{(16)}$ & 0.0042$^{(11)}$  & 157 & 819 & 976\\
21 & Daniel2018 & 0.1736$^{(18)}$ & 0.2172$^{(26)}$ & 0.0994$^{(16)}$ & 0.0723$^{(33)}$ & 0.1115$^{(38)}$ & 0.0075$^{(26)}$  & 158 & 330 & 488\\
22 & simonss & 0.1745$^{(19)}$ & 0.2043$^{(20)}$ & 0.1196$^{(22)}$ & 0.0462$^{(16)}$ & 0.0682$^{(15)}$ & 0.0043$^{(12)}$  & 157 & 1164 & 1321\\
23 & hihi123 & 0.1767$^{(20)}$ & 0.2106$^{(23)}$ & 0.1133$^{(21)}$ & 0.0456$^{(14)}$ & 0.0730$^{(19)}$ & 0.0042$^{(11)}$  & 184 & 644 & 828\\
24 & Sungmin & 0.1780$^{(21)}$ & 0.2086$^{(22)}$ & 0.1210$^{(24)}$ & 0.0473$^{(18)}$ & 0.0697$^{(17)}$ & 0.0044$^{(13)}$  & 136 & 1160 & 1296\\
25 & HIT\_face & 0.1780$^{(21)}$ & 0.2083$^{(21)}$ & 0.1224$^{(25)}$ & 0.0470$^{(17)}$ & 0.0701$^{(18)}$ & 0.0041$^{(10)}$  & 226 & 639 & 865\\
26 & thesherlock & 0.1826$^{(22)}$ & 0.2124$^{(24)}$ & 0.1285$^{(28)}$ & 0.0501$^{(21)}$ & 0.0744$^{(21)}$ & 0.0045$^{(14)}$  & 162 & 439 & 601\\
27 & fh\_nj & 0.1827$^{(23)}$ & 0.2144$^{(25)}$ & 0.1249$^{(26)}$ & 0.0508$^{(22)}$ & 0.0758$^{(22)}$ & 0.0048$^{(16)}$  & 158 & 316 & 474\\
28 & liu\_xiang886 & 0.1859$^{(24)}$ & 0.2329$^{(32)}$ & 0.0897$^{(14)}$ & 0.1103$^{(46)}$ & 0.1543$^{(48)}$ & 0.0127$^{(33)}$  & 157 & 317 & 474\\
29 & ppnn & 0.1867$^{(25)}$ & 0.2251$^{(29)}$ & 0.1056$^{(19)}$ & 0.0937$^{(44)}$ & 0.1282$^{(43)}$ & 0.0133$^{(34)}$  & 159 & 827 & 986\\
30 & MCPRL\_aiwa & 0.1882$^{(26)}$ & 0.2182$^{(27)}$ & 0.1328$^{(30)}$ & 0.0500$^{(20)}$ & 0.0741$^{(20)}$& 0.0045$^{(14)}$  & 158 & 657 & 815\\
31 & tongtong & 0.1923$^{(27)}$ & 0.2246$^{(28)}$ & 0.1332$^{(31)}$ & 0.0529$^{(23)}$ & 0.0792$^{(25)}$& 0.0047$^{(15)}$  & 159 & 318 & 477\\
32 & billzeng & 0.1923$^{(27)}$ & 0.2246$^{(28)}$ & 0.1332$^{(31)}$ & 0.0529$^{(23)}$ & 0.0792$^{(25)}$& 0.0048$^{(16)}$  & 157 & 319 & 476\\
33 & runauto & 0.1938$^{(28)}$ & 0.2311$^{(31)}$ & 0.1224$^{(25)}$ & 0.0741$^{(34)}$ & 0.1066$^{(35)}$& 0.0071$^{(24)}$  & 157 & 733 & 890\\
34 & yossibiton & 0.1944$^{(29)}$ & 0.2263$^{(30)}$ & 0.1345$^{(32)}$ & 0.0552$^{(25)}$ & 0.0818$^{(27)}$& 0.0049$^{(17)}$  & 167 & 83 & 250\\
35 & haoyayu365 & 0.1970$^{(30)}$ & 0.2345$^{(33)}$ & 0.1321$^{(29)}$ & 0.0566$^{(26)}$ & 0.0892$^{(28)}$& 0.0061$^{(19)}$  & 156 & 396 & 552\\
36 & dler & 0.2064$^{(31)}$ & 0.2389$^{(34)}$ & 0.1464$^{(41)}$ & 0.0538$^{(24)}$ & 0.0807$^{(26)}$ & 0.0051$^{(18)}$  & 181 & 1060 & 1241\\
37 & f.gomes & 0.2092$^{(32)}$ & 0.2459$^{(36)}$ & 0.1407$^{(34)}$ & 0.0676$^{(30)}$ & 0.0999$^{(32)}$ & 0.0064$^{(21)}$  & 157 & 1285 & 1442\\
38 & AntonS & 0.2093$^{(33)}$ & 0.2450$^{(35)}$ & 0.1411$^{(35)}$ & 0.0698$^{(31)}$ & 0.1017$^{(33)}$ & 0.0069$^{(23)}$  & 157 & 337 & 494\\
39 & Jim\_1021 & 0.2106$^{(34)}$ & 0.2553$^{(40)}$ & 0.1204$^{(23)}$ & 0.0978$^{(45)}$ & 0.1384$^{(46)}$ & 0.0101$^{(31)}$  & 156 & 829 & 985\\
40 & linghu8812 & 0.2117$^{(35)}$ & 0.2510$^{(38)}$ & 0.1352$^{(33)}$ & 0.0828$^{(39)}$ & 0.1200$^{(41)}$ & 0.0072$^{(25)}$  & 157 & 857 & 1014\\
41 & tib6913 & 0.2141$^{(36)}$ & 0.2501$^{(37)}$ & 0.1453$^{(39)}$ & 0.0720$^{(32)}$ & 0.1044$^{(34)}$ & 0.0066$^{(22)}$  & 157 & 192 & 349\\
42 & meixitu2 & 0.2171$^{(37)}$ & 0.2501$^{(37)}$ & 0.1529$^{(46)}$ & 0.0630$^{(28)}$ & 0.0901$^{(29)}$ & 0.0069$^{(23)}$  & 157 & 295 & 452\\
43 & Cavall & 0.2174$^{(38)}$ & 0.2583$^{(44)}$ & 0.1418$^{(37)}$ & 0.0792$^{(37)}$ & 0.1172$^{(39)}$ & 0.0066$^{(22)}$  & 157 & 389 & 546\\
44 & HYL\_Dave & 0.2174$^{(38)}$ & 0.2562$^{(41)}$ & 0.1444$^{(38)}$ & 0.0751$^{(36)}$ & 0.1112$^{(37)}$ & 0.0069$^{(23)}$  & 157 & 173 & 330\\
45 & zhangge00hou & 0.2187$^{(39)}$ & 0.2565$^{(42)}$ & 0.1469$^{(42)}$ & 0.0742$^{(35)}$ & 0.1090$^{(36)}$ & 0.0071$^{(24)}$  & 157 & 647 & 804\\
46 & maguih & 0.2188$^{(40)}$ & 0.2566$^{(43)}$ & 0.1469$^{(42)}$ & 0.0742$^{(35)}$ & 0.1090$^{(36)}$ & 0.0071$^{(24)}$  & 158 & 651 & 809\\
47 & jinhong.zhang & 0.2202$^{(41)}$ & 0.2539$^{(39)}$ & 0.1524$^{(45)}$ & 0.0661$^{(29)}$ & 0.0934$^{(31)}$ & 0.0071$^{(24)}$  & 12 & 341 & 353\\
48 & cmkyec & 0.2286$^{(42)}$ & 0.2781$^{(46)}$ & 0.1413$^{(36)}$ & 0.0932$^{(43)}$ & 0.1419$^{(47)}$ & 0.0103$^{(32)}$  & 158 & 247 & 405\\
49 & nikkonew & 0.2295$^{(43)}$ & 0.2681$^{(45)}$ & 0.1545$^{(47)}$ & 0.0816$^{(38)}$ & 0.1185$^{(40)}$ & 0.0077$^{(27)}$  & 156 & 321 & 477\\
50 & litian1045 & 0.2342$^{(44)}$ & 0.2816$^{(48)}$ & 0.1483$^{(44)}$ & 0.0871$^{(40)}$ & 0.1341$^{(45)}$ & 0.0094$^{(29)}$  & 157 & 172 & 329\\
\hline
\end{tabular}}}
\end{center}
\vspace{-4mm}
\caption{The leaderboard of first phase. Results outperforming the baseline (Participant: \emph{litian1045}) are shown.}
\vspace{-3mm}
\label{table:competition_results}
\end{table*}

\subsection{Competition Ranking}

The competition is ranked according to both MFR and SFR metrics. To reduce a tendency that models overfit on masked or standard face recognition, the main series of evaluation metrics are designed to show a weighted sum to consider both masked and standard faces at the same time.
 As shown in Table \ref{table:competition_results}, the overall ranking is ascend ordered by the \emph{All (MFR\&SFR)} metric: \emph{All (MFR\&SFR)} = 0.25 $\times$ \emph{All-Masked} + 0.75 $\times$ \emph{All (SFR)}. At the same time, \emph{Wild} and \emph{Controlled} metrics can also be computed as:
\emph{Wild (MFR\&SFR)} = 0.25 $\times$ \emph{Wild-Masked} + 0.75 $\times$ \emph{Wild (SFR)};
\emph{Controlled (MFR\&SFR)} = 0.25 $\times$ \emph{Controlled-Masked} + 0.75 $\times$ \emph{Controlled (SFR)}. It is worth noting that scores of different metrics is computed at corresponding FNMR@FMR=10-5.

\begin{figure*}
\small
\centering
\subfigure[]{
\includegraphics[width=0.32\linewidth]{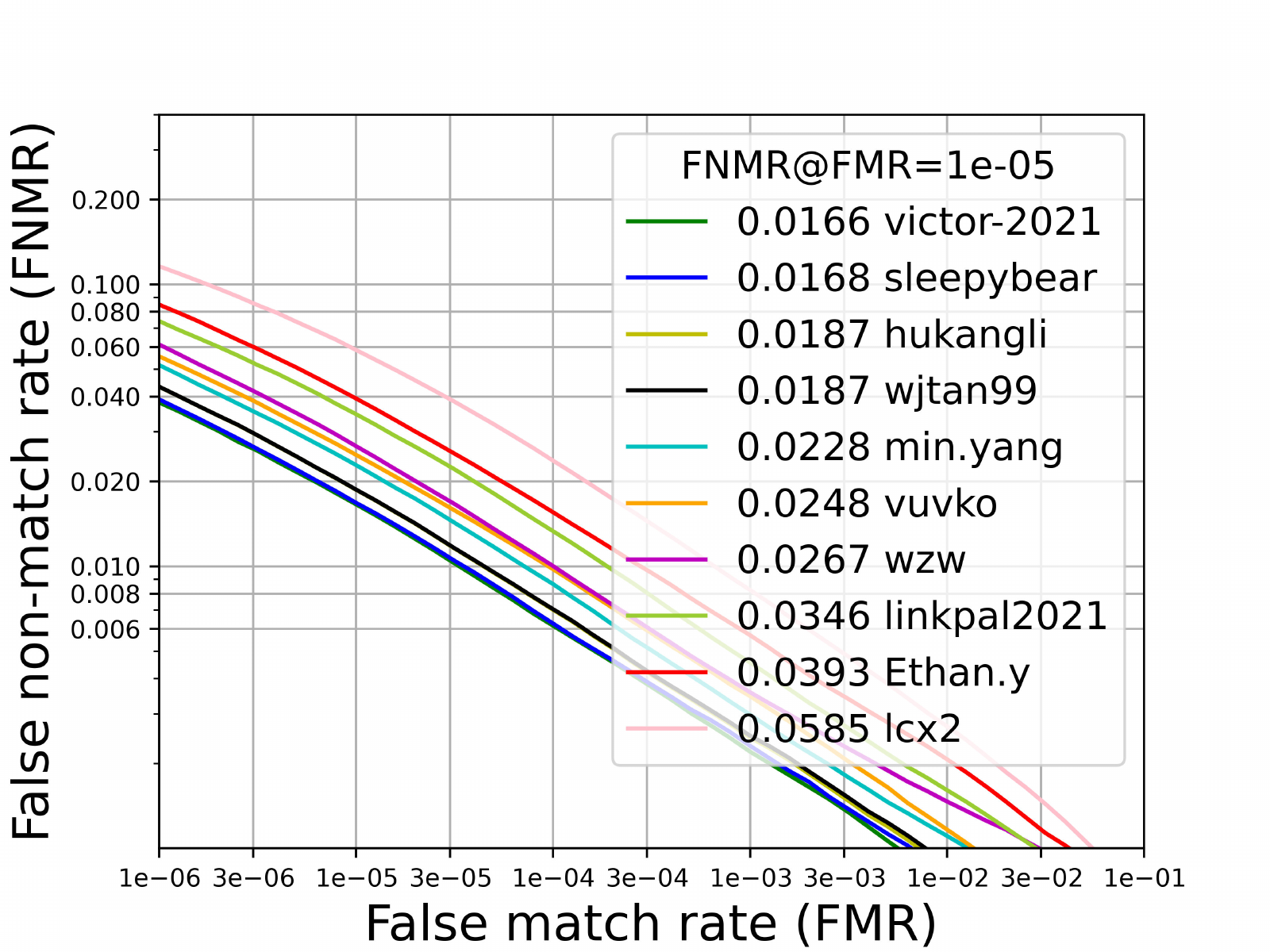}}
\subfigure[]{
\includegraphics[width=0.32\linewidth]{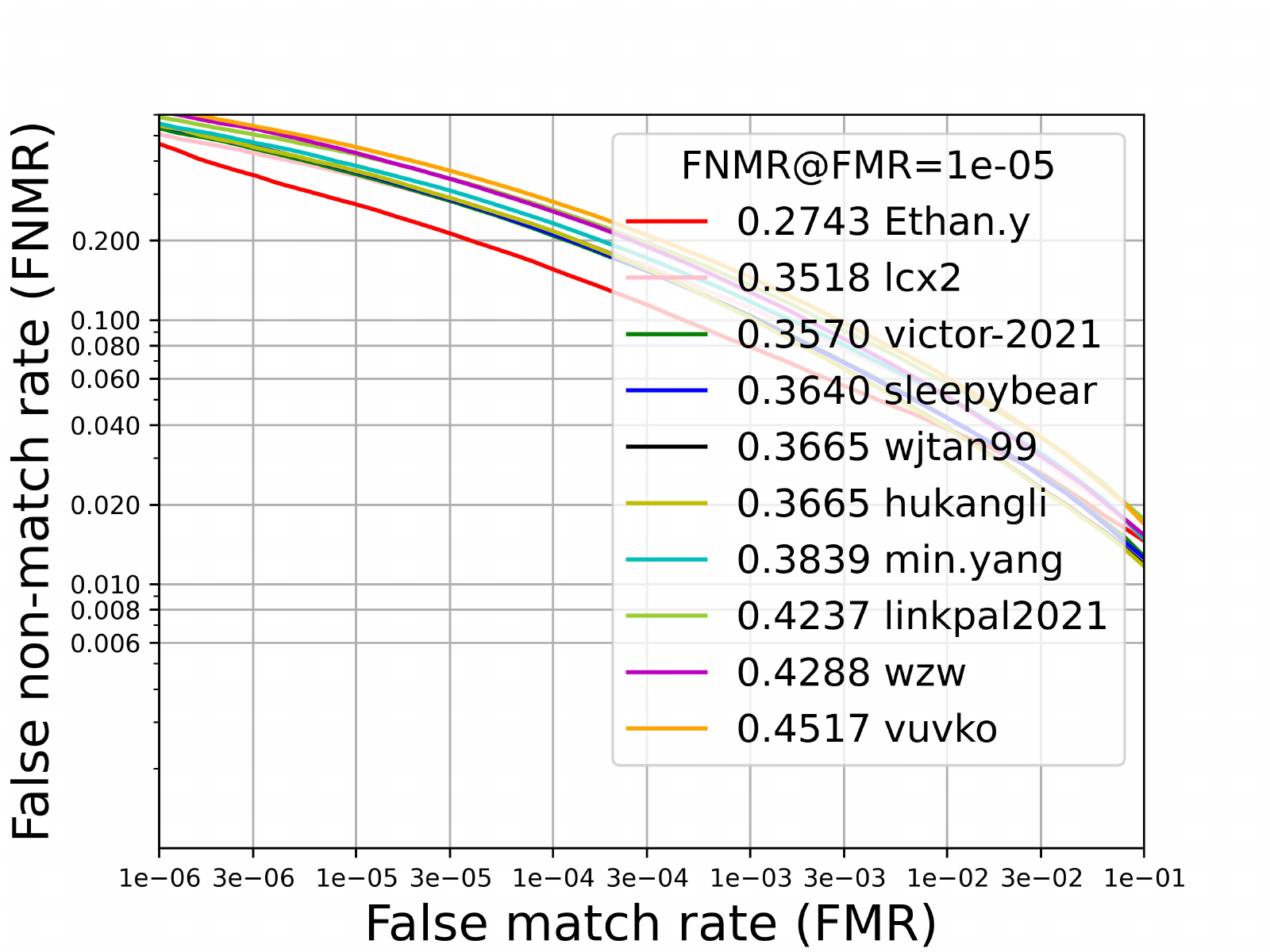}}
\subfigure[]{
\includegraphics[width=0.32\linewidth]{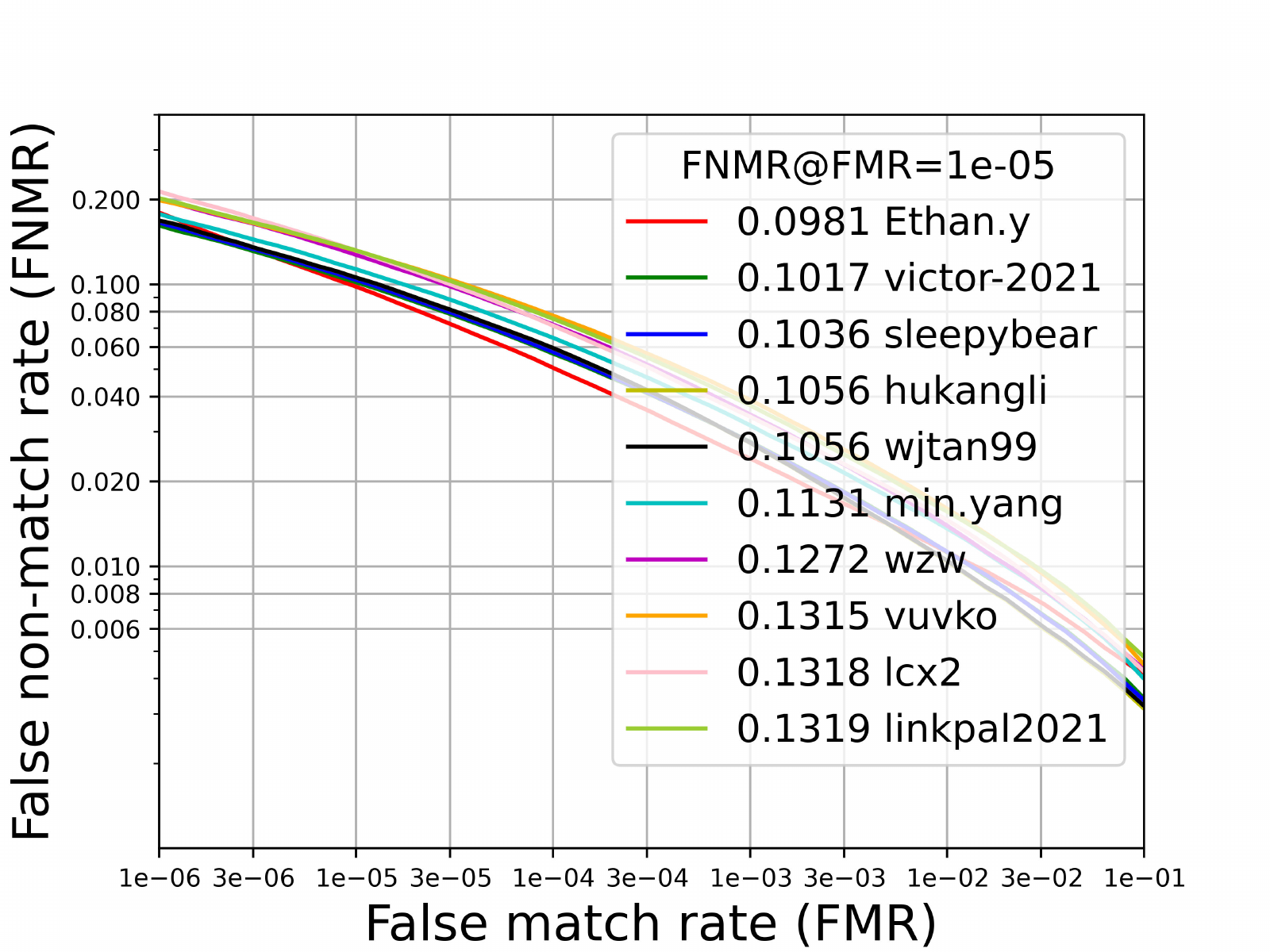}}
\vspace{-3mm}
\caption{FMR-FNMR plots of SFR, MFR, and final combined MFR\&SFR results.}
\vspace{-3mm}
\label{fig:mask_test_set}
\end{figure*}

\section{Baseline Solutions}

\subsection{Implementation Details}

In order to fairly evaluate the performance of different face recognition models, we reproduce representative algorithms in one Gluon codebase with the hyper-parameters referred to the original papers. Default batch size per GPU is set as 64 unless otherwise indicated. Learning rate is set as 0.05 for a single node (8 GPUs), and follows the linear scaling rule \cite{goyal2017accurate} for the training on multiple nodes (\ie $0.05\times$\# machines). We decrease the learning rate by 0.1$\times$ at 8, 12, and 16 epochs, and stop at 20 epochs for all models. During training, we only adopt the flip data augmentation. Note that other data augmentations such as adding simulated mask are encouraged to boost MFR performance.

\subsection{Baseline Model and Results}

The configuration of the baseline model is ResNet-50 backbone, ArcFace loss, with WebFace12M (30\%) training data. The backbone architecture is shown in Table~\ref{table:backbone}.
The evaluation results of the baseline model (Participant: \emph{litian1045}) are shown in the last row of Table~\ref{table:competition_results}. The FNMR@FMR=1e-5 across different attributes (\emph{All (MFR\&SFR)}, \emph{Wild (MFR\&SFR)}, \emph{Controlled (MFR\&SFR)}, \emph{All (SFR)}, \emph{Wild (SFR)}, \emph{Controlled (SFR)}) is 0.2342, 0.2816, 0.1483, 0.0871, 0.1341, 0.0094, respectively.
The time cost of detection, recognition and total is 157 ms, 172 ms, 329 ms, respectively.

\begin{table}[!t]
\begin{center}{\scalebox{0.9}{
\begin{tabular}{c|c|c}
\hline
layer name & 50-layer & output size \\
\hline\hline
Input Image Crop &  & 112$\times$112$\times$3\\\hline
 & 3$\times$3, 64, stride~1 & 112$\times$112$\times$64\\\hline
Conv2\_x &  $\begin{bmatrix}\makecell[c]{3\times3, 64\\3\times3, 64}\end{bmatrix}\times3$  & $56\times56\times64$\\\hline
Conv3\_x &  $\begin{bmatrix}\makecell[c]{3\times3, 128\\3\times3, 128}\end{bmatrix}\times4$  & $28\times28\times128$\\\hline
Conv4\_x &  $\begin{bmatrix}\makecell[c]{3\times3, 256\\3\times3, 256}\end{bmatrix}\times14$  & $14\times14\times256$\\\hline
Conv5\_x &  $\begin{bmatrix}\makecell[c]{3\times3, 512\\3\times3, 512}\end{bmatrix}\times3$  & $7\times7\times512$\\\hline
FC & & 512\\\hline
\end{tabular}}}
\end{center}
\vspace{-2mm}
\caption{The network configuration of our baseline model.
Convolutional building blocks are shown in brackets with the numbers of blocks stacked. Down-sampling is performed by the second conv in conv2\_1, conv3\_1, conv4\_1, and conv5\_1 with a stride of 2.}
\vspace{-2mm}
\label{table:backbone}
\end{table}

\section{Preliminary Results of First Phase}


First phase was held between June 7, 2021 to August 11, 2021. During this phase, 69 teams submit a total of 833 effective solutions to the challenge. Table~\ref{table:competition_results} gives a detailed result for each participant with their ranks. As shown in Table~\ref{table:competition_results} and Figure \ref{fig:mask_test_set}, \emph{Ethan.y} ranks the 1st place in main \emph{All (MFR\&SFR)} metric with 0.0980 as well as \emph{Controlled (MFR\&SFR)} with 0.0500, while \emph{victor-2021} wins \emph{Wild (MFR\&SFR)} with a score of 0.1222.
For SFR metrics, \emph{victor-2021} ranks the 1st place among \emph{All (SFR)}, \emph{Wild (SFR)} and \emph{Controlled (SFR)} with 0.0162, 0.0270 and 0.0018. Since the FNMR of MFR is much higher than that in SFR, the MFR performance dominates the final ranks.
It is worth noting that during first phase, a model is allowed to be evaluated if its \emph{Total time} (which is a sum of \emph{Detection time} and \emph{Recognition time}) is no more than 2000ms. In the final ranks (second phase), we only rank models with a \emph{Total time} less than 1000ms. As shown in Table~\ref{table:competition_results}, there are 9 participants who have submitted models to the leaderboard with \emph{Total time} more than 1000ms.
Since the challenge is still going on, more details of top-ranked solution would be updated in the future.


\section{Conclusion}

To address the MFR problem during epidemic, we organize the Face Bio-metrics under COVID Workshop and Masked Face Recognition Challenge in ICCV 2021.
Enabled by the WebFace260M and FRUITS, this challenge (WebFace260M Track) aims to push the frontiers of practical MFR.
This report details the training data, evaluation protocols, submission rules, test set and metric in SFR and MFR, ranking criterion, baseline solution, and preliminary competition results.
In the first phase of WebFace260M Track, 69 teams (total 833 solutions) participate in the challenge and 49 teams exceed the performance of our baseline.
We will actively update this report in the future.

{\small
\bibliographystyle{ieee_fullname}
\bibliography{egbib}
}

\end{document}